\pdfoutput=1
\documentclass[11pt]{article}
\usepackage[utf8]{inputenc}
\usepackage[T1]{fontenc}
\usepackage[]{acl}
\usepackage{times}
\usepackage{latexsym}
\usepackage{microtype}
\usepackage{booktabs} 
\usepackage{color}
\usepackage{amsmath}
\usepackage{amsfonts}
\usepackage{graphicx}
\usepackage{bbm}
\usepackage{multirow, multicol}
\usepackage{comment}
\usepackage{xspace}
\usepackage{upquote}
\usepackage{float}
\usepackage{alltt}
\usepackage{algpseudocode}
\usepackage{enumitem}
\usepackage{mathpartir}
\usepackage{wrapfig}
\usepackage{array}
\usepackage{pifont}

\usepackage[ruled, vlined]{algorithm2e} 

\SetAlFnt{\small}
\SetAlCapFnt{\small}
\SetAlCapNameFnt{\small}
\SetAlCapHSkip{0pt}
\SetArgSty{text}
\SetCommentSty{text}
\IncMargin{-\parindent}
\DontPrintSemicolon

\definecolor{comment-gray}{rgb}{0.5, 0.5, 0.5}

\SetCommentSty{mycommfont}

\definecolor{comment-red}{rgb}{0.8,0,0}

\newcommand{\Agent}[0]{\text{Agent}\xspace}
\newcommand{\Sim}[0]{\text{Sim}\xspace}
\newcommand{\Syn}[0]{\text{Syn}\xspace}
\newcommand{\Deploy}[0]{\text{Deploy}\xspace}
\newcommand{\CSP}[0]{\text{CSP}\xspace}
\newcommand{\D}[0]{\mathcal{D}\xspace}
\newcommand{\M}[0]{\text{Sim}\xspace}
\newcommand{\dTT}[0]{ThingTalk\xspace}
\newcommand{\ins}[0]{in-simulation\xspace}
\newcommand{\oos}[0]{out-of-simulation\xspace}
\newcommand{\DM}[0]{\D_\M\xspace}

\newcommand{\cmark}{\ding{51}}%

\newcommand{\ctxnlu}[0]{CSP\xspace}

\newcommand{\skimnoagent}[0]{{\small\textsc{CSP-NoAgent}}\xspace}

\title{A Few-Shot Semantic Parser for Wizard-of-Oz Dialogues \\
with the Precise ThingTalk Representation}

\author{Giovanni Campagna \quad Sina J. Semnani \quad Ryan Kearns \quad Lucas Jun Koba Sato\\
\quad \textbf{Silei Xu} \quad \textbf{Monica S. Lam} \\
  Computer Science Department \\
  Stanford University \\
  Stanford, CA, USA \\
  \texttt{\{gcampagn,sinaj,kearns,satojk,silei,lam\}@cs.stanford.edu}}

\date{}

\begin{document}
\maketitle
\begin{abstract}
Previous attempts to build effective semantic parsers for Wizard-of-Oz (WOZ) conversations suffer from the difficulty in acquiring a high-quality, manually annotated training set. Approaches based only on dialogue synthesis are insufficient, as dialogues generated from state-machine based models are poor approximations of real-life conversations. Furthermore, previously proposed dialogue state representations are ambiguous and lack the precision necessary for building an effective agent.

This paper proposes a new dialogue representation and a sample-efficient methodology that can predict precise dialogue states in WOZ conversations. We extended the ThingTalk representation to capture all information an agent needs to respond properly. Our training strategy is sample-efficient: we combine (1) few-shot data sparsely sampling the full dialogue space and (2) synthesized data covering a subset space of dialogues generated by a succinct state-based dialogue model. The completeness of the extended \dTT language is demonstrated with a fully operational agent, which is also used in training data synthesis. 

We demonstrate the effectiveness of our methodology on MultiWOZ 3.0, a reannotation of the MultiWOZ 2.1 dataset in \dTT. \dTT can represent 98\% of the test turns, while the simulator can emulate 85\% of the validation set. We train a contextual semantic parser using our strategy, and obtain 79\% turn-by-turn exact match accuracy on the reannotated test set.\footnote{Our data and code can be downloaded from \url{https://oval.cs.stanford.edu/releases/}}

\end{abstract}

\section{Introduction}

\begin{figure}[t]
\centering
\includegraphics[width=0.9\linewidth]{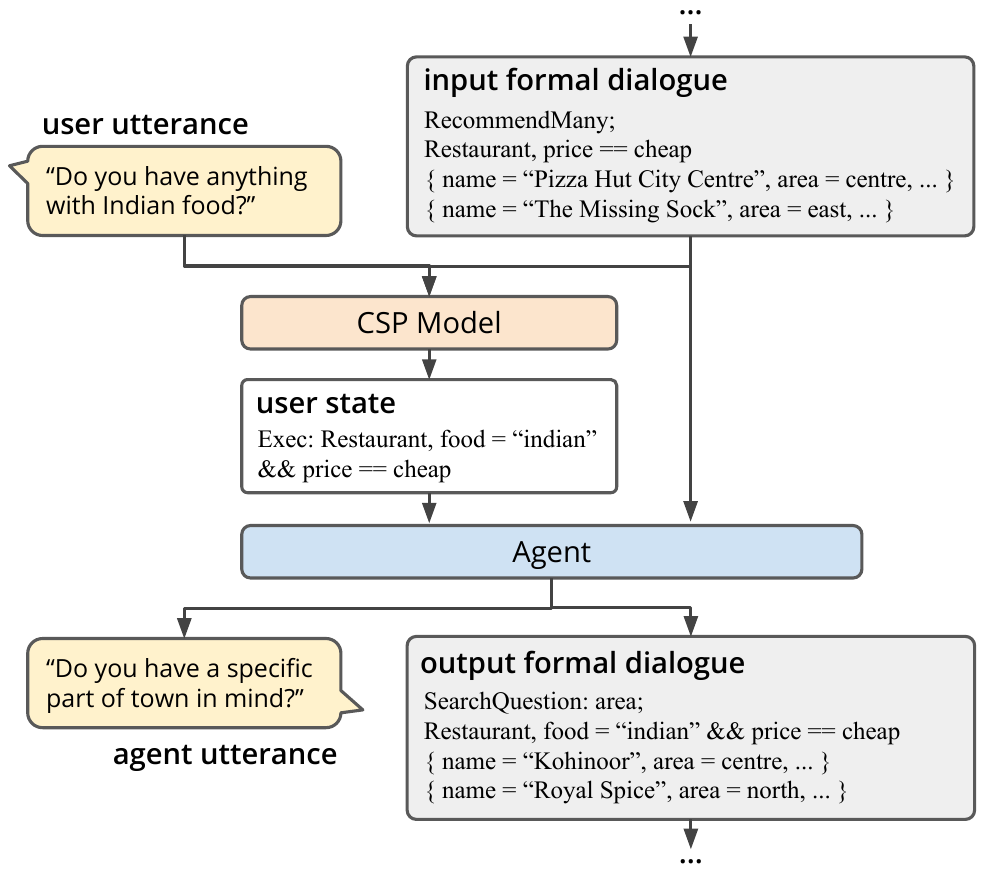}
\caption{The inference-time flow of a dialogue agent with a contextual semantic parser based on the \dTT representation.}
\label{fig:dialogue-loop}
\vspace{-1em}
\end{figure}

Virtual assistants and task-oriented dialogue agents are transforming how consumers interact with computers. This has led to active research on dialogue state tracking networks~\cite{ren2019scalable, zhou2019multi, zhang2019task, chenschema, heck2020trippy}, and even full neural networks that track dialogue states, implement dialogue policies, and generate agent utterances~\cite{williams2016end, eric2017key, zhang2019task, peng2020soloist, hosseini2020simple}.

Dialogue state tracking on Wizard-of-Oz task-oriented conversations, where humans are asked to simulate both the agent and the user, has proven to be challenging. For example, despite multiple rounds of manual annotation, the MultiWOZ multi-domain task-oriented dataset still contains significant errors which hamper the development of accurate semantic parsers \cite{zang2020multiwoz, han2020multiwoz, ye2021multiwoz}. An approach to bypass manual annotations is to generate dialogues using a simulator and then manually paraphrase them~\cite{DBLP:journals/corr/abs-1801-04871}. Unfortunately, as we shall show in this paper, 
such dialogue simulators do not exercise many of the possible dialogue flows seen in Wizard-of-Oz conversations. This gap is likely to widen with real-life conversations. 

Given the many attempts to create accurate semantic parsers for the MultiWOZ data set, this paper takes a fresh look at the problem of understanding Wizard-of-Oz conversations.  We observe two fundamental flaws with the current approach. 
Previously proposed state representations such as slot-value pairs and the recently proposed hierarchical forms~\cite{cheng-etal-2020-conversational} do not capture critical details in the user utterances, such as logical ``or'' and negation.
Even if the semantic parser is 100\% accurate, the agent will not be able to satisfy the user's request.  Second, it is easy to make errors. The existing slot representation is ambiguous, so it is not possible to be consistently correct. This leads to poor quality of annotation. 

This paper shows that it is possible to create a precise and accurate semantic parser for Wizard-of-Oz conversations in a sample-efficient manner. We introduce the MultiWOZ 3.0 dataset, a reannotation of the full test set and partial validation set of MultiWOZ 2.1~\cite{eric2019multiwoz}, using a new, more precise formal representation. 
The contributions of this paper include:

{\bf 1. A precise, complete, executable ThingTalk representation for dialogues.} In previous work, we proposed the ThingTalk programming language to represent just a single utterance~\cite{geniepldi19}. Here we extend it to a full formal representation of a dialogue, including multiple turns of user input, results from the user request (such as a database lookup or API invocation), and the agent's response. We show that the extended ThingTalk for dialogues is precise enough to capture 98\% of the turns in MultiWOZ 3.0. In the rest of the paper, we will refer to the extended ThingTalk language as ThingTalk, unless noted otherwise. 

We also demonstrate that ThingTalk is a complete representation for dialogues. The agent directly executes the ThingTalk representation to retrieve the results from the databases and APIs, without referring to any of the user utterances.
In fact, the same agent code can be used both during simulation and in a real agent deployment.

{\bf 2. We show that we can obtain a high-quality synthetic training data set with a simulator that adopts the ThingTalk representation.} The precision of ThingTalk makes it possible to generate many  distinctively different dialogue paths that mirror those in the WOZ conversation.  Our experiment shows that our simulator can generate 85\% of the user turns. 

{\bf 3. We show that by leveraging synthesized dialogues represented in ThingTalk, we can train an effective semantic parser for WOZ conversations.} This is significant since it is difficult to annotate dialogues accurately.  ThingTalk does not make it easier to annotate, but it is unambiguous. We annotate manually only a \textit{few-shot} training set, and rely on synthesis for the rest. The few-shot training data is 2\% of the typical amount of annotated data. 

The few-shot training samples in ThingTalk help the semantic parser generalize from the simulated dialogues to WOZ conversations. Whereas the simulator can only generate a subset of the states representable by ThingTalk, ThingTalk can precisely represent nearly all WOZ data.

Our novel contextual semantic parser, described in Section~\ref{sec:training}, obtains a turn-by-turn accuracy of 79\% on MultiWOZ 3.0. Note that this model generalizes to utterances that fall out of the realm of simulation. 

\section{Related Work}

\paragraph{State Representation for DST}
\textit{Dialogue State Tracking} is the task of predicting a formal representation of a conversation. 
The standard representation used in DST contains the values of all slots mentioned in the dialogue~\cite{asri2017frames, budzianowski2018large}. This is inadequate in practice. First of all, the definition is ambiguous, as it could mean ``all slots mentioned by the user'' or ``all slots mentioned by either the user or the agent''. This has lead to inconsistency in the annotation. Second, the representation does not track the comparison or logical operators in the request, so it cannot model complex queries.

Recently, \newcite{cheng-etal-2020-conversational} proposed adopting a formal representation for both the user and agent state, using the TreeDST representation.
TreeDST was built to support only dialogues synthesized and paraphrased from a compatible state machine, while \dTT supports the full generality of Wizard-of-Oz conversations. 

\paragraph{Data Acquisition for DST}
In recent years, a number of very large DST datasets have been released~\cite{budzianowski2018large,byrne2019taskmaster,rastogi2019towards}. The preferred technique to acquire such datasets is through Wizard-of-Oz~\cite{kelley1984iterative}, a technique in which two humans are instructed to converse with each other, with one person taking the role of the agent. WOZ datasets are expensive, and the annotation quality is poor.
A different approach synthesizes a large corpus of dialogues using a state machine, then employs crowdworkers to paraphrase them. Paraphrasing has been applied to semantic parsing~\cite{overnight} and dialogues~\cite{DBLP:journals/corr/abs-1801-04871,rastogi2019towards,cheng-etal-2020-conversational}. Paraphrased datasets have less variety than WOZ, and crowdsourced paraphrases are also expensive. 
Our approach has a significant cost advantage, while matching the variety of WOZ dialogues.

\newcite{zeroshotmultiwoz} found that using data synthesized from a small finite state machine, it is possible to increase the accuracy of DST in the transfer learning setting. Later, \newcite{yu2020score} proposed using synthesized data to pre-train a DST model, using a different objective function. They showed modest improvements in MultiWOZ 2.1, using the full training set. We instead propose using the same fine-tuning objective for both synthesized and few-shot annotated data.

\section{The \dTT Dialogue Language}

The \dTT Dialogue Language is designed to formally capture all relevant information in task-oriented dialogues to interpret what the user says next. This includes the user utterances, the result of the user requests, as well as the agent's replies.  

To see why the results and the agent's reply are needed, consider the example in Fig.~\ref{fig:dialogue-loop}. The user has previously asked for a cheap restaurant, and now asks ``Do you have anything with Indian food?''. In the example, the agent noted that there are many cheap restaurants available, so it is likely that the user wants both ``Indian'' and ``cheap''. This is reflected in the query that the command maps to. Conversely, had the agent responded that there are no cheap restaurants, it is likely that the user no longer cares about finding a cheap and only wants Indian. The user query thus would be just:
\begin{tabbing}
123\=\kill
$\text{Exec} : \text{Restaurant}, \textit{food}=\text{``indian''}$
\end{tabbing}
This illustrates that the meaning of the user utterance depends on the result and the agent's response, so we must include them in the formal dialogue.
The previous slot-based representation captures only what is mentioned by the user; it is not precise enough to handle this example. 

\begin{figure}
\small
\begin{tabbing}
123456\=1\=\kill
{\bf (a) Sorting and ranking in \dTT}\\
\>Agent:\'There are 14 trains that arrive by 12:45. What time \\
\>would you like to leave?\\
\>User:\'What's the latest train i can take that will still get me\\
\>there by 12:45?\\
\>$u_1 =$\'$\text{Exec}: \texttt{sort}(\textit{arrive\_by} ~\texttt{desc}~~\texttt{of}~~\textit{Train},$\\
\>\>$\textit{arrive\_by} \le \text{12:45} ~~\texttt{\&\&}~~ \ldots)[1]$\\
\\
{\bf (b) Projection and logical operators in \dTT}\\
\>User:\'I think i would like to visit both churchill and \\
\>magdalene colleges. May I have their phone \\ \>numbers?\\
\>$u_1 =$\'$\text{Exec}: [\textit{phone}] ~~\texttt{of}~~ \text{Attraction},$\\
\>\>$\textit{name}=\text{``churchill''} ~~\vert\vert~~\textit{name}=\text{``magdalene''}$
\end{tabbing}
\vspace{-1em}
\caption{\dTT representations of user utterance examples in the MultiWOZ 3.0 validation set. $u_1$ denotes the user state.}
\vspace{-0.5em}
\label{fig:query-examples}
\end{figure}

Formally, \dTT represents (1) the user state $u \in U$ with the semantics of a single user turn, (2) the agent state $a \in A$ with the semantics of the single agent turn, and (3) the formal dialogue $d \in \D$ to capture all information necessary to interpret the user utterance. In this section, we provide the detailed definition of each component. The formal syntax is included in Appendix \ref{sec:syntax}.

\paragraph{User State.}
The formal semantics of a user turn is represented by a \textit{user state} $u \in U$, which consists of an abstract dialogue act and, for dialogue acts that provide or request information, a sequence of \textit{statements}: either database queries, or actions with side effects (such as making a reservation). Queries specify the domain of interest and can use the standard relational operators: selection, projection, aggregation, sorting. Actions specify the domain, the action name, and the parameters necessary for the action. User state examples in Figures \ref{fig:dialogue-loop} and \ref{fig:query-examples} 
with abstract act ``Exec'' are all queries, while the example in Fig.~\ref{fig:incomplete-example} uses the action ``Restaurant.MakeReservation''.

The user state includes new statements that are implied by the current utterance and statements that the user has previously mentioned and is still interested in pursuing (Fig.~\ref{fig:incomplete-example}). 
Note that a single user utterance may map to multiple \dTT statements, possibly in different domains.

\begin{figure}
\small
\begin{tabbing}
123456\=1\=12345\=\kill
\>Agent:\'[\ldots] Would you like me to make you a reservation?\\
\>User:\'Yes, please make a reservation.\\
\>$u_1 =$\'$\text{Exec}: \text{Restaurant}.\text{MakeReservation}(\textit{name}=\text{``...''})$\\
\>Agent:\'What day and time?\\
\>$a_1 = $\'SlotFill: \textit{book\_day}, \textit{book\_time}\\
\>$\text{Restaurant}.\text{MakeReservation}(\textit{name}=\text{``...''})$\\
{\bf (a) User answers the question}\\
\>User:\' At 17:30 on Friday.\\
\>$u_2 =$\'$\text{Exec}: \text{Restaurant}.\text{MakeReservation}(\textit{name}=\text{``...''},$\\ \>\>$\textit{book\_time}=\text{17:30}, \textit{book\_day}=\text{friday});$\\
{\bf (b) Or, user switches to a new domain instead}\\
\>User:\'Nevermind. Not at this time. Can you help me find \\
\>the postcode for the Holiday Inn Cambridge?\\
\>$u_2 =$\'$\text{Exec}: \text{Hotel}, \textit{name}=\text{``holiday inn cambridge''};$
\end{tabbing}
\vspace{-1em}
\caption{Examples of a user  continuing or abandoning a transaction, adapted from the MultiWOZ 3.0 validation set. The user state $u_2$ denotes this fact by propagating or discarding the action. $a_1$ is the agent state.}
\vspace{-1em}
\label{fig:incomplete-example}
\end{figure}

\paragraph{Agent State.}
Analogously, each agent turn has a formal \textit{agent state} $a \in A$ representation, which is computed by the agent policy. The agent state includes an abstract dialogue act, as well as an optional agent statement, which either requests some slots from the user, proposes a new statement to the user, or asks the user to confirm an action.

\paragraph{Formal Dialogue Representation.}
A formal dialogue $d\in \D$ captures all the information in the conversation needed to interpret the user utterance. Specifically, it contains the current agent state, the accumulated results of executing the user statements in previous turns, and the user statements that the user has asked to execute but that are missing some required parameters. The results for queries are the items retrieved from the database; the results for actions are returned by the API call.

\begingroup
\setlength{\tabcolsep}{2pt}
\begin{table}
\small
\centering
\begin{tabular}{p{3.5cm}cccc}
\toprule
\bf Feature & Slots & TreeDST & Express & TT \\
\midrule
\multicolumn{4}{l}{\bf User} \\
Executable Semantics     & $\times$ & $\times$ & {\bf \cmark} & {\bf \cmark} \\
Canonicalizable     & $\times$ & $\times$ & $\times$ & {\bf \cmark} \\
Greetings       & $\times$ & $\times$ & ? & {\bf \cmark} \\
Learn More, Ask Recomm. & $\times$ & $\times$ & ? & {\bf \cmark} \\
Multi-domain Turns       & $\times$ & {\bf \cmark} & {\bf \cmark} & {\bf \cmark} \\
Request Features: & & & \\
\quad Slot Constraints & {\bf \cmark} & {\bf \cmark} & {\bf \cmark} & {\bf \cmark} \\
\quad Comparisons     & $\times$ & {\bf \cmark} & {\bf \cmark} & {\bf \cmark} \\
\quad Logical And & {\bf \cmark} & {\bf \cmark} & {\bf \cmark} & {\bf \cmark} \\
\quad Logical Or, Not & $\times$ & $\times$ & {\bf \cmark} & {\bf \cmark} \\
\quad Projection & $\times$ & {\bf \cmark} & {\bf \cmark} & {\bf \cmark} \\
\quad Ranking & $\times$ & $\times$ & ? & {\bf \cmark}\\
\midrule
\multicolumn{4}{l}{\bf Agent} \\
Dialogue Acts   & $\times$ & {\bf \cmark} & {\bf \cmark} & {\bf \cmark} \\
Requested Slots & {\bf \cmark} & {\bf \cmark} & {\bf \cmark} & {\bf \cmark} \\
Proposed Slots & $\times$ & {\bf \cmark} & {\bf \cmark} & {\bf \cmark} \\
\bottomrule
\end{tabular}
\caption{Comparison of representation power for different lexical features of different formal dialogue languages. TreeDST refers to \newcite{cheng-etal-2020-conversational}, Express refers to \newcite{SMDataflow2020}. TT indicates \dTT.}
\label{table:thingtalk-comparison}
\vspace{-1em}
\end{table}

\paragraph{Comparison with previous representations}

In Table~\ref{table:thingtalk-comparison} we compare \dTT with three existing state representation: the slots and values representation used in MultiWOZ, the TreeDST representation~\cite{cheng-etal-2020-conversational}, and the Express representation~\cite{SMDataflow2020, smdataflow2021}. Note that neither Express nor TreeDST are open-source or available to use, whereas ThingTalk is fully open-source and comes with tools that developers can use. Limited documentation exists for Express, so we use ``?'' for features we do not know are supported or not.
    
\dTT represents user queries and commands as {\em executable} database queries and API calls. An executable representation is easier to annotate manually. Other approaches require annotators to be familiar with the semantics of each domain, whereas in our approach annotators just need to learn the database query syntax to annotate for different domains. Additionally, the implementation of the agent only needs to execute \dTT statements; no custom per-domain logic is necessary. 

Furthermore, \dTT is \textit{canonicalizable}: the annotation of the semantics of a turn is syntactically unique, regardless of how the turn is phrased, and the unique form can be computed automatically. This is important both to enforce conventions on manually annotated data, as well as to be able to paraphrase: if the annotation depends on the syntactic form of the utterance, the annotation must be changed after paraphrasing. Express, while executable, is not canonicalizable because it represents coreferences explicitly and expresses updates to the dialogue state as edits. Both features lead to syntactically different representations for the same semantics, for example if the coreference is by name, by constraints, or by pronoun.

\dTT can represent the full generality of WoZ conversations. For example, \dTT can represent turns that have no request, at the beginning and end of the conversation. Neither slots nor TreeDST have a representation for those turns. This oversight highlights the need to design the representation based on real conversations.

One feature present in the previous representation that we drop from \dTT is the precise slots mentioned by the agent.  For example, in response to a user asking for a restaurant, the agent may mention the restaurant ``name'' and ''address.''
Such slots do not affect the interpretation of the user utterance. Removing them from the agent state coalesces many more utterances into the same state, and allows to approximate more complex human agent utterances, increasing the state coverage and boosting the accuracy of the semantic parser.
\section{Simulator-Agent Architecture}
To synthesize data for training, we propose a \textit{simulator-agent} architecture. The state-based simulator takes the role of the human user. The same agent that would be used at deployment time is used during synthesis. The agent is built based on the semantics of \dTT, not just the simulator. It can respond correctly to any dialogue $d \in \D$ representable in \dTT. On the other hand, the simulator samples a subset space $\DM \subset \D$. We refer to dialogues in $\DM$ as \textit{\ins}; other dialogues are \textit{\oos}.

Formally, the architecture has three components: 
\begin{description}
\item
$\Agent(d, u):\D \times U \rightarrow \D$: an agent that accepts a formal dialogue $d \in \D$, and the user state $u \in U$ representing the last user utterance, to produce a new dialogue $d' \in \D$. The agent guarantees that if $d \in \DM$ then $d' \in \DM$.

\item $\Sim(d): \DM \rightarrow X \times U$: a simulator that accepts an \ins dialogue $d \in \DM$, and non-deterministically creates a new user utterance $x \in X$ and its user state $u \in U$.

\item $\CSP(d,x): \D \times X \rightarrow U$, a contextual semantic parsing model that accepts a dialogue $d \in \D$, which may not be in $\DM$, and a user utterance $x \in X$ to predict the user state in $U$. 
\end{description}

In this section, we describe how the components are used to synthesize training data and build a functional dialogue agent. 

\subsection{Training Data Synthesis}
We synthesize training data for CSP as follows:
\begin{description}
\item
$\Syn(d): \DM \rightarrow \DM \times X \times U$: the synthesizer accepts a dialogue $d \in \DM$ and returns a training sample produced by using \Sim to generate a possible user utterance and a resulting \ins dialogue to be predicted, then applying the \Agent to continue: \begin{align*}
\Syn(d) &= (d', x, u), \text{ where }\\
(x, u) &= \Sim(d), d' = \Agent(d, u)
\end{align*}
\end{description}
Starting with a null dialogue, we iteratively use $\Syn$ to synthesize training samples. During synthesis, the agent is called in a mock execution environment with no side effects, and it uses a non-deterministic policy that generates many possible agent behaviors. It is helpful to include many agent behaviors because it helps model the human WOZ agent. 

Following \newcite{zeroshotmultiwoz}, both the simulator and the agent policy are implemented using a domain-independent state machine which includes many natural language templates for user and agent utterances. Using the templates and a few natural language phrases for each slot, we can generate dialogues for any new domain with minimal effort.

\subsection{Deployment}
After training, the same agent can be used at deploy time to reply to the real user.
\begin{description}
\item
$\Deploy(d,x): \D \times X \rightarrow \D$: given the current dialogue, a deployable system uses \CSP to map the next user utterance to a formal dialogue, which is then used by \Agent to continue the dialogue. 
Let $d_0$ be the empty dialogue and user input ${x_1,x_2,\dots}$
\vspace{-0.5em}
\begin{align*}
d_{i} &= \Deploy(d_{i-1}, x_i) \\
      &= \Agent(d_{i-1}, \CSP(d_{i-1}, x_i))
\end{align*}
\end{description}

\subsection{Out-of-simulation Dialogues}

\begin{figure}
\small
\begin{tabbing}
123456\=1\=\kill
\>User:\'Please book a table for 5 at 14:30 on wednesday at  \\
\> Royal Spice. I also need to find a place to stay.\\
\>$u_1=$\'$\text{Exec}: \text{Restaurant}.\text{MakeReservation}($\\
\>\>$\textit{name}=\text{``royal spice''}, \textit{book\_people}=5,$\\
\>\>$\textit{book\_time}=\text{14:30}, \textit{book\_day}=\text{wednesday});$\\
\>\>$\text{Hotel};$\\
\>Agent:\'I was able to book your table successfully.\\
\>Your reference number is kqmxil0z. Now, what \\
\>type of accommodations are you looking for today?\\
\end{tabbing}
\vspace{-1.5em}
\caption{Example of an out-of-simulation dialogue from the MultiWOZ 3.0 test set, where the same turn mentions two domains. The simulator never generates such a turn but the agent can reply to it.}
\label{fig:intro-examples}
\vspace{-1em}
\end{figure}

While the simulator can cover only the most common dialogue paths, \dTT is designed to be general, covering many more possible dialogues. To improve generality, the CSP is trained not only with simulated dialogues but also few-shot data annotated with the full expressiveness of \dTT. Correspondingly, the agent is written to handle the full representation of \dTT. This design makes our parser and agent more robust than those that only train with simulated dialogues. 
Fig.~\ref{fig:intro-examples} shows an out-of-simulation dialogue from the MultiWOZ test set. In the example, the agent must reply to two domains at once.

We show below some of the \oos dialogue patterns handled by our agent. 
\begin{itemize}[noitemsep,topsep=0pt]
\item \textit{Domain switch}: the user switches to a new domain in the middle of a discussion about another; the simulator switches domains only after completing the action.

\item \textit{Multidomain}: the user refers to two domains in the same utterance; the simulator only refers to one domain at a time.

\item \textit{Eager action parameters}:
the user specifies parameters for an action before completing the query, ignoring a prompt from the agent to refine the query.

\item \textit{Abandoning transactions}: the user abandons a transaction after it has been initiated; the simulator never interrupts a transaction.  
\end{itemize}
These examples illustrate the many plausible ways in which the user can change the course of a dialogue. Trying to simulate all these possibilities is infeasible, nor is it desirable, as it will worsen the distribution of the training data by overemphasizing uncommon patterns. At the same time, handling these cases is important; thus, we train with few-shot annotated data and rely on the model's inherent generalization capability.

\section{Contextual Semantic Parsing Model}
\label{sec:training}

\subsection{Model Architecture}
\label{sec:model}

Our CSP neural model is fine-tuned from the pre-trained BART model~\cite{lewis2019bart}. BART is a Transformer encoder-decoder neural network~\cite{vaswani2017attention} pre-trained with the task of reconstructing noised inputs. 
Our model for the user encodes a concatenation of the formal dialogue and the user utterance, and is trained to generate the user state as its output. 

To reduce the length of the input, the formal dialogue is truncated before feeding to the model: only the last executed query and action in each domain are kept, and the rest is discarded. Previous statements are no longer relevant; information that is still relevant is carried over in the last statement. 
Additionally, we encode at most one result per query. We observe that the user uses either a coreference to refer to the only/first choice, or uses the entity name. The model is trained to copy entity names from the user utterance.

We use BART-Large, with about 400M parameters. We train it with token-level cross-entropy loss and teacher forcing. Hyperparameters and preprocessing details are included in Appendix \ref{sec:hyperparameter}.

\subsection{Training Data}

\paragraph{Data Synthesis.}
We use \Syn to synthesize an initial set of training dialogues, covering all possible combinations of slots at each turn, and many possible paths in $\DM$.

\paragraph{Automatic Paraphrasing.}

We apply \emph{automatic paraphrasing with filtering}~\cite{xu2020autoqa} to increase the variety of natural language in each turn. We use a pre-trained BART model fine-tuned on the ParaBank2 general-purpose paraphrasing dataset~\cite{parabank2}. Each user utterance is paraphrased individually. We apply \textit{filtering} to ensure that the user state does not change for each utterance: each paraphrased utterance, with its associated formal dialogue, is passed to a model trained on synthesized data; the utterance is discarded if the model predicts a different user state than the annotation before paraphrasing.

\paragraph{Few-Shot Fine-Tuning.}
To expose the model to the variety in real-world data, we fine-tune the model with a small number of manually annotated dialogues. 

\paragraph{Self-Training.}

Acquiring large fully-annotated WOZ datasets is challenging, because annotations are often erroneous. 
Acquiring \textit{unannotated} WOZ datasets, on the other hand, is easier.
To use such data when available, we propose using \textit{self-training} \cite{mcclosky2006effective, einolghozati2019improving, zoph2020rethinking}.  
We apply the model fine-tuned on few-shot data to unannotated input, create a training set using the predicted result as annotations, and use that to further fine-tune the model. 

The annotation of WOZ dialogues requires predictions of the agent state as well, unlike the simulated dialogues where the agent state is generated automatically. We apply the same methodology as for the user states to the agent state, so as to annotate the full dialogues for training.

\section{Evaluation}

Our evaluation attempts to answer these research questions:
\begin{enumerate}[noitemsep,topsep=0pt,leftmargin=*]
\item How well does our \dTT representation model Wizard-of-Oz conversations?
\item What accuracy can a model achieve in the task of predicting \dTT, given our training data acquisition strategy?
\item How well do our dialogue simulator and our dialogue agent approximate real dialogues? 
\end{enumerate}

\subsection{Experimental Setting}
We conduct our experiments using the MultiWOZ dataset~\cite{budzianowski2018large, eric2019multiwoz}. This dataset includes English task-oriented dialogues across five domains: Attraction, Hotel, Restaurant, Taxi, and Train.

We reannotated parts of MultiWOZ 2.1 with \dTT annotations, and we name this version MultiWOZ 3.0. The authors of this paper reannotated the full test set and, due to a lack of time, 36\% of the validation set, discarding the rest. Our result is thus a lower-bound on the possible accuracy: with more of the validation set annotated, we expect higher test accuracy.

The slot values in our new test set differ from the original annotations in 83\% of the turns. This is not surprising because 
others have already found problems in MultiWOZ 2.1~\cite{zhou2019multi, zang2020multiwoz, han2020multiwoz}, and because \dTT and the existing annotations adopt different conventions for when a slot should be included. We found mistakes in the annotations, inconsistent normalization of names, and inconsistent annotation of slots offered by the agent. We dropped 1\% of test turns due to unrecoverable human errors, such as the user acting as the agent. 

We use four datasets for training:
\begin{itemize}[noitemsep,topsep=0pt]
\item 
{\em Synthesized} dataset, generated using our state-machine-based simulator and agent, consisting of around 1M dialogues across all five domains. The state machine has 20 abstract transitions for the agent, and 43 for the user.
\item
{\em Paraphrase} dataset, obtained by automatically paraphrasing the synthesized data. 
\item
\textit{Few-Shot} dataset, a split of 168 dialogues from the original validation set. This amounts to 2\% of the original training set. Another 265 dialogues in the original validation set are used as the 3.0 validation set.
\item
\textit{Self-Trained} dataset, obtained by self-training on the MultiWOZ training set. 
\end{itemize}
Dataset statistics are detailed in Appendix~\ref{sec:data-stats}.

We use the Genie Toolkit~\cite{geniepldi19} for data synthesis and Hugging Face's Transformers library~\cite{Wolf2019HuggingFacesTS} for the model.

\subsection{Precision of \dTT}
\label{eval-dTT}
\dTT is designed to precisely cover the semantics of Wizard-of-Oz dialogues.   We first observe that \dTT captures the semantics of the sentences well: it can represent the validation set in its entirety, and 99.8\% of the user utterances and 97.6\% of the agent utterances in the test set are representable. Overall, that comprises 97.7\% of the test turns. \dTT cannot represent, for example, out-of-domain questions, questions that cannot be answered using the given database, 
 and agent utterances such as asking the users to wait.

User utterances in the test set that cannot be represented are simply counted as errors, while agent utterances that cannot be represented as marked with a single ``invalid'' dialogue act, which is given as input to the neural model. The model can choose to ignore the invalid dialogue act and attempt to predict the correct user state regardless.

\subsection{Accuracy on the MultiWOZ 3.0 Test Set}

Our first experiment evaluates how well our \ctxnlu model can understand the user utterances in the MultiWOZ 3.0 dataset on four metrics. \\
{\bf Exact match accuracy} requires the predicted user state to identically match the annotation. \\
{\bf Slot accuracy} requires the slots provided by the user in the predicted user state to match the annotation, ignoring comparison operators, requested slots, and the dialogue act. \\ 
{\bf Turn-by-turn accuracy} assumes that the gold dialogue up to the current turn is available as input.\\
{\bf Dialogue accuracy} requires predicting the correct state for {\em all} the previous and current turns of a given dialogue. This is a challenging but meaningful metric because in practice, once the model fails, the conversation diverges from the WOZ dialogue. 

We train our \ctxnlu model on the combination of Synthesized and Paraphrased sets, fine-tune it on the Few-Shot training set, and fine-tune it again on the Self-Trained set. Our model achieves a 79.2\% turn-by-turn accuracy and 44.1\% dialogue accuracy in exact match (Table~\ref{table:main-results}).

To understand the role of synthesized data, we removed all synthesized data, and train with only the manually annotated few-shot data. The synthesized data improves the turn-by-turn exact match accuracy by 5.5\% and the dialogue exact match accuracy by 8.4\%. This shows that the low-cost automatically generated training data is effective.

We performed an ablation study on the validation set to evaluate the components of our training strategy  (Table~\ref{table:main-results}). We first observe that the validation accuracy is higher than the test accuracy, because we used the validation set to refine our synthesis.  Training with only synthesized data already delivers a respectable 61.8\% turn-by-turn accuracy; with the augmentation of auto-paraphrasing data, turn-by-turn accuracy improves 0.1\%, and dialogue accuracy improves 0.4\%. 

The few-shot training alone delivers a high accuracy of 75.6\%.  When the model trained on synthesized and paraphrased data is fine-tuned with few-shot data, the accuracy is 81.0\%, showing that these two approaches complement each other. Self-training further improves the turn-by-turn accuracy by 0.4\%, with 1\% better dialogue accuracy.

\begin{table}
\setlength{\tabcolsep}{4.4pt}
\small
\centering
\begin{tabular}{llrrrr}
\toprule
\multicolumn{2}{l}{\multirow{2}{*}{\bf Training Strategy}} & \multicolumn{2}{c}{\bf Turn-by-Turn} & \multicolumn{2}{c}{\bf Dialogue} \\
& & \multicolumn{1}{c}{EM} & \multicolumn{1}{c}{Slot} & \multicolumn{1}{c}{EM} & \multicolumn{1}{c}{Slot} \\
\midrule
\multirow{2}{*}{Test} & Full training  & \bf 79.2\% & \bf 87.5\% & \bf 44.1\% & \bf 61.0\% \\
& Few-shot only  & 73.7\% & 81.6\% & 35.7\% & 46.3\% \\
\midrule
\multirow{5}{*}{Dev} &
Full training        & \bf 81.4\% & \bf 88.7\% & \bf 51.9\% & \bf 67.2\% \\        
&$-$ self-training                & 81.0\% & 88.0\% & 50.9\% & 65.3\% \\
&Synth. only             &     61.8\% & 73.1\% & 29.1\% & 38.0\% \\
&Synth. + para.              &     61.9\% & 73.3\% & 29.5\% & 37.4\% \\
&Few-shot only           &     75.6\% & 81.7\% & 41.8\% & 51.6\% \\
\bottomrule
\end{tabular}
\caption{Turn-by-turn and dialogue accuracy, both exact match (EM) and slot, of the \ctxnlu model, on the MultiWOZ 3.0 test and validation sets.}
\label{table:main-results}
\end{table}

\subsection{Generalization of the Dialogue Model}
Our strategy is to handle the complexity of Wizard-of-Oz dialogues with a combination of simulated dialogues and few-shot training samples to teach generalization beyond simulated dialogues. We analyze the validation set to understand the difference between the simulated dialogues and the Wizard-of-Oz dialogues, and its effect on accuracy.

The results are shown in Table~\ref{table:synthesis-eval}. The validation set is divided into:
\begin{enumerate}[noitemsep,topsep=0pt]
\item \text{Trained}: 15.5\% of the validation set turns  share the same formal dialogue and user state with some sample in training (ignoring the slot values). Accuracy obtained: 93.1\%.

\item \text{In-simulation}: 69.7\% of the validation set turns can be represented by the simulator: the formal context is contained in $\DM$, and the user state can be generated by the simulator. Accuracy obtained: 82.4\%.

\item \text{Out-of-simulation}: 14.7\% of the validation turns require the model to generalize beyond $\DM$, either through few-shot or its own generalization capabilities. Accuracy obtained: 62.7\%.
\end{enumerate}

\begin{table}
\small
\centering
\begin{tabular}{lrr}
\toprule
{\bf Category}   & {\bf \% Turns} & {\bf Accuracy} \\
\midrule
Trained          & 15.5\%         & 93.1\%         \\
In-simulation    & 69.7\%         & 82.4\%         \\
\hline
Out-of-simulation  & 14.7\%         & 62.7\%         \\
\quad Unknown agent state  & 6.3\%         & 66.0\%         \\
\quad Domain switch  & 4.0\%         & 84.0\%         \\
\quad Eager action parameters  & 0.9\%         & 73.3\%         \\
\quad Multidomain          & 0.8\%         & 16.7\% \\
\quad Abandon transaction  & 0.5\%         & 25.0\%         \\
\bottomrule
\end{tabular}
\caption{Turn-by-turn exact match accuracy of validation set, categorized by whether each user utterance is synthesizable by our simulator. For the unsynthesizable category, we further divide in common classes of user behavior not captured by the simulator.}
\label{table:synthesis-eval}
\vspace{-1em}
\end{table}

{\bf Our synthesizer covers the Wizard-of-Oz conversations well.} Even though our simulator and agent are built using a state machine with only 54 user transitions and 24 agent transitions, 85.2\% of the validation set is in-simulation.

{\bf Research that trains and validates on simulated data is missing a non-trivial population of Wizard-of-Oz dialogues.}
We found that 14.7\% of the validation turns are representable in \dTT but are out-of-simulation. 

{\bf Our training strategy generalizes beyond the simulated dialogues.}
For the out-of-simulation turns, our model achieves an accuracy of 62.7\%. 
The model can generalize well on validation turns where the agent state is unseen in training, achieving 66\% accuracy. This result speaks to the strength of using a formal representation of the agent, which avoids interpreting untrained agent utterances. 

The model also reacts well to strong signals in the user utterance. The model achieves 84.1\% accuracy when the user switches domains unexpectedly, and 73.3\% accuracy when the user starts issuing slots for the action before completing the query.
 
Finally, when the user issues a command over two domains at once, the model achieves 16.7\% accuracy. When the user abandons a booking transaction mid-way, the model achieves 25\% accuracy. These kinds of out-of-simulation states are also rare in the few shot training set. The model can generalize, but is biased towards the common cases seen in the training data.

\subsection{Dialogue History vs. Formal Context}
\begin{table}
\small
\centering
\begin{tabular}{p{2.1cm}lrr}
\toprule
{\bf Model} & {\bf Training Data} & {\bf Accuracy} \\
\midrule
TRADE & MultiWOZ 2.1 &  37.3\%    \\
TRADE & 0-shot 2.1  & 12.1\%    \\
SUMBT & MultiWOZ 2.1 &  39.3\%    \\
SUMBT & 0-shot 2.1  & 18.3\%    \\
STAR & MultiWOZ 2.1  & \bf 49.9\%    \\
\mbox{\skimnoagent} & MultiWOZ 2.1 & 45.6\% \\
\mbox{\skimnoagent}       & 0-shot 2.1 &  13.3\%    \\
\mbox{\skimnoagent}       & $+$ auto-parap. & 12.2\%    \\
CSP & MultiWOZ 3.0 & 37.3\% \\
CSP & Synthesized & 23.6\% \\
CSP & $+$ auto-parap. & 25.2\% \\
\bottomrule
\end{tabular}
\caption{Dialogue slot accuracy on the MultiWOZ 2.1 test set. \skimnoagent has no formal agent state; it encodes the previous slots, and the current agent and user utterances. 0-shot 2.1 is the synthesized data by~\newcite{zeroshotmultiwoz}. CSP was trained on MultiWOZ 3.0 but tested on 2.1.}
\label{table:up-to-error-comparison}
\vspace{-1em}
\end{table}

We wish to evaluate the difference between using dialogue history, as in DST models, and using a formal context. We do so by measuring the dialogue accuracy, which has the same definition for DST and CSP. 

Because we do not have the resources to reannotate the training data with \dTT, we will use the MultiWOZ 2.1 training set for this experiment.
For a DST parser, we use TRADE~\cite{Wu2019May}, SUMBT~\cite{Lee2019Jul}, and STAR~\cite{ye2021star}, three high-performing models for MultiWOZ 2.1. For CSP, we train a model we call \skimnoagent, which uses the same neural architecture as our CSP. Because MultiWOZ 2.1 has no formal agent state annotations, \skimnoagent uses the original slot-value annotation from the immediately preceding turn as the formal input context. This context, the current agent utterance, and the current user utterance are used to predict all the slots from the dialogue. This is the best approximation to \dTT possible given the available data; the results provide a lower bound on CSP with fully annotated training data. 

The results are shown in Table~\ref{table:up-to-error-comparison}. We see that \skimnoagent outperforms TRADE by 8.3\% and SUMBT by 6.3\% in dialogue accuracy, and is within 4\% of STAR, a highly optimized model. Note that \skimnoagent needs {\em no new annotations}, and the slot representation captures only a small subset of the information in the utterances. This shows the advantage of replacing the dialogue history with a formal context. It also shows that the use of formal contexts can be applied in other representations.

For comparison, we also test our CSP on MultiWOZ 2.1, using self-predicted formal agent states. Our model, trained on MultiWOZ 3.0, reaches 37.3\% dialogue accuracy in the MultiWOZ 2.1 test set. 
This is due to the reannotation of MultiWOZ 3.0, and because the model is trained and tested on data with different annotation conventions. Compared to the dialogue slot accuracy on MultiWOZ 3.0, we observe a gap of about 11\%, which serves as a lower bound on the benefit of having experts annotate the test data. Note that our approach does not require manual annotation of a large training set, and therefore expert annotation of test data was feasible.

\subsection{Comparison with Previous 0-Shot Model}
Our last experiment compares our work with the zero-shot model proposed by \newcite{zeroshotmultiwoz}. Their paper only included results with transfer learning on new domains. Here, we evaluate TRADE, SUMBT, and \skimnoagent trained with their synthesized data in a zero-shot fashion.  The results shown in Table~\ref{table:up-to-error-comparison} indicate that the previous approach is inadequate, achieving only 12.1\% dialogue accuracy with TRADE and 18.3\% with SUMBT.  \skimnoagent achieves 13.3\% dialogue accuracy. 
Our approach, instead, achieves 23.6\% dialogue accuracy. Adding automatic paraphrasing increases the turn-by-turn accuracy by about 3\% for both models.

This result shows that our approach is much more effective in synthesizing data. In particular, it is important to represent the agent state formally when training with synthesized data, as it eliminates the need to synthesize and parse agent utterances. 

\section{Conclusion}

This paper presents a sample-efficient methodology, based on the extended \dTT representation, to predict precise dialogue states in Wizard-of-Oz conversations. We achieve a turn-by-turn exact-match accuracy of 79.2\% on the MultiWOZ 3.0 dataset, while using 50x less manually annotated training data than the original MultiWOZ dataset.

The proposed  \dTT dialogue representation is precise, complete, and executable. It is {\em precise} enough to cover 98\% of the dialogue turns in MultiWOZ.  The precision enables automatic synthesis of dialogues covering 85\% of the MultiWOZ data set.  \dTT is {\em complete} and {\em executable}, as evidenced by a fully working agent that can simply execute  \dTT queries without referring to the user input. Furthermore, the agent can handle dialogue flows beyond those that can be simulated. 

The accuracy is achieved with a contextual semantic parser (CSP) where the dialogue context is represented in  \dTT rather than  the natural language dialogue history. It is trained first with
auto-paraphrased synthetic data, fine-tuned with the few-shot annotated data, then self-trained.  

In summary, this paper shows that with  \dTT, we can predict WOZ dialogues accurately with training data mostly generated from a state machine. Our methodology thus combines the best of the WOZ and M2M approaches, as it can handle the more realistic WOZ dialogues, while having a low data acquisition cost like M2M. 

\section{Ethical Considerations}
We envision that our training strategy will broaden the availability of task-oriented agents for tasks and populations not currently covered by existing large-scale datasets, due to its
low annotation requirement. We have open-sourced tool set designed around our representation for bootstrapping affordable contextual semantic parsers for new domains. 

Our agent was tuned and evaluated on the MultiWOZ benchmark. MultiWOZ is a crowdsourced Wizard-of-Oz dataset; WOZ datasets are known not to fully represent real-world conversations~\cite{ganhotra-etal-2020-effects}. Further research is needed before a dialogue agent based on our methodology can be deployed in the real world. Additionally, the current version of the agent was tuned for English; future work should investigate techniques to automatically localize a contextual semantic parser, analogous to prior research done for single-turn semantic parsers~\cite{moradshahi-etal-2020-localizing}.

Our training strategy replaces manual annotation of data with automatically obtained data, which requires some additional amount of computation time. In practice, such additional compute is small: data synthesis runs in 5 hours on a single machine with no GPUs; the paraphrase dataset can be obtained in about 5 hours on a machine with 4 Nvidia T4 GPUs; training completes within 8 hours on a machine with one Nvidia V100; self-training requires 2 hours on a single Nvidia T4 GPU, and fine-tuning is another 1.5 hours on one Nvidia V100. Overall, the whole process is done with about 22 hours of compute time, well below the cost of human annotation of equivalent amounts of data. We note that the large amount of synthetic data poses no challenge to convergence in practice, so training models with a large amount of synthesized data has little effect on the compute cost.

The manually annotated portion of our dataset was obtained from the previously released MultiWOZ 2.1 dataset, a crowdsourced dataset. No crowdsourcing was employed in this paper; the data was annotated by the authors. 

\section*{Acknowledgments}
This work is supported by the National Science Foundation
under Grant No. 1900638, and the Alfred P. Sloan Foundation under Grant No. G-2020-13938.

\bibliographystyle{acl_natbib}
\bibliography{paper}

\begin{thebibliography}{39}
\expandafter\ifx\csname natexlab\endcsname\relax\def\natexlab#1{#1}\fi

\bibitem[{Andreas et~al.(2020)Andreas, Bufe, Burkett, Chen, Clausman, Crawford,
  Crim, DeLoach, Dorner, Eisner, Fang, Guo, Hall, Hayes, Hill, Ho, Iwaszuk,
  Jha, Klein, Krishnamurthy, Lanman, Liang, Lin, Lintsbakh, McGovern,
  Nisnevich, Pauls, Petters, Read, Roth, Roy, Rusak, Short, Slomin, Snyder,
  Striplin, Su, Tellman, Thomson, Vorobev, Witoszko, Wolfe, Wray, Zhang, and
  Zotov}]{SMDataflow2020}
Jacob Andreas, John Bufe, David Burkett, Charles Chen, Josh Clausman, Jean
  Crawford, Kate Crim, Jordan DeLoach, Leah Dorner, Jason Eisner, Hao Fang,
  Alan Guo, David Hall, Kristin Hayes, Kellie Hill, Diana Ho, Wendy Iwaszuk,
  Smriti Jha, Dan Klein, Jayant Krishnamurthy, Theo Lanman, Percy Liang,
  Christopher~H. Lin, Ilya Lintsbakh, Andy McGovern, Aleksandr Nisnevich, Adam
  Pauls, Dmitrij Petters, Brent Read, Dan Roth, Subhro Roy, Jesse Rusak, Beth
  Short, Div Slomin, Ben Snyder, Stephon Striplin, Yu~Su, Zachary Tellman, Sam
  Thomson, Andrei Vorobev, Izabela Witoszko, Jason Wolfe, Abby Wray, Yuchen
  Zhang, and Alexander Zotov. 2020.
\newblock \href {https://doi.org/10.1162/tacl_a_00333} {Task-oriented dialogue
  as dataflow synthesis}.
\newblock \emph{Transactions of the Association for Computational Linguistics},
  8:556--571.

\bibitem[{Budzianowski et~al.(2018)Budzianowski, Wen, Tseng, Casanueva, Stefan,
  Osman, and Ga{\v{s}}i\'c}]{budzianowski2018large}
Pawe{\l} Budzianowski, Tsung-Hsien Wen, Bo-Hsiang Tseng, I{\~n}igo Casanueva,
  Ultes Stefan, Ramadan Osman, and Milica Ga{\v{s}}i\'c. 2018.
\newblock Multi{WOZ} - a large-scale multi-domain wizard-of-oz dataset for
  task-oriented dialogue modelling.
\newblock In \emph{Proceedings of the 2018 Conference on Empirical Methods in
  Natural Language Processing (EMNLP)}.

\bibitem[{Byrne et~al.(2019)Byrne, Krishnamoorthi, Sankar, Neelakantan,
  Goodrich, Duckworth, Yavuz, Dubey, Kim, and Cedilnik}]{byrne2019taskmaster}
Bill Byrne, Karthik Krishnamoorthi, Chinnadhurai Sankar, Arvind Neelakantan,
  Ben Goodrich, Daniel Duckworth, Semih Yavuz, Amit Dubey, Kyu-Young Kim, and
  Andy Cedilnik. 2019.
\newblock Taskmaster-1: Toward a realistic and diverse dialog dataset.
\newblock In \emph{Proceedings of the 2019 Conference on Empirical Methods in
  Natural Language Processing and the 9th International Joint Conference on
  Natural Language Processing (EMNLP-IJCNLP)}, pages 4506--4517.

\bibitem[{Campagna et~al.(2020)Campagna, Foryciarz, Moradshahi, and
  Lam}]{zeroshotmultiwoz}
Giovanni Campagna, Agata Foryciarz, Mehrad Moradshahi, and Monica Lam. 2020.
\newblock \href {https://doi.org/10.18653/v1/2020.acl-main.12} {Zero-shot
  transfer learning with synthesized data for multi-domain dialogue state
  tracking}.
\newblock In \emph{Proceedings of the 58th Annual Meeting of the Association
  for Computational Linguistics}, pages 122--132, Online. Association for
  Computational Linguistics.

\bibitem[{Campagna et~al.(2019)Campagna, Xu, Moradshahi, Socher, and
  Lam}]{geniepldi19}
Giovanni Campagna, Silei Xu, Mehrad Moradshahi, Richard Socher, and Monica~S.
  Lam. 2019.
\newblock \href {https://doi.org/10.1145/3314221.3314594} {Genie: A generator
  of natural language semantic parsers for virtual assistant commands}.
\newblock In \emph{Proceedings of the 40th ACM SIGPLAN Conference on
  Programming Language Design and Implementation}, PLDI 2019, pages 394--410,
  New York, NY, USA. ACM.

\bibitem[{Chen et~al.(2020)Chen, Lv, Wang, Zhu, Tan, and Yu}]{chenschema}
Lu~Chen, Boer Lv, Chi Wang, Su~Zhu, Bowen Tan, and Kai Yu. 2020.
\newblock Schema-guided multi-domain dialogue state tracking with graph
  attention neural networks.
\newblock In \emph{Proceedings of the AAAI Conference on Artificial
  Intelligence}, volume~34, pages 7521--7528.

\bibitem[{Cheng et~al.(2020)Cheng, Agrawal, Mart{\'\i}nez~Alonso, Bhargava,
  Driesen, Flego, Kaplan, Kartsaklis, Li, Piraviperumal, Williams, Yu,
  {\'O}~S{\'e}aghdha, and Johannsen}]{cheng-etal-2020-conversational}
Jianpeng Cheng, Devang Agrawal, H{\'e}ctor Mart{\'\i}nez~Alonso, Shruti
  Bhargava, Joris Driesen, Federico Flego, Dain Kaplan, Dimitri Kartsaklis, Lin
  Li, Dhivya Piraviperumal, Jason~D. Williams, Hong Yu, Diarmuid
  {\'O}~S{\'e}aghdha, and Anders Johannsen. 2020.
\newblock \href {https://www.aclweb.org/anthology/2020.emnlp-main.651}
  {Conversational semantic parsing for dialog state tracking}.
\newblock In \emph{Proceedings of the 2020 Conference on Empirical Methods in
  Natural Language Processing (EMNLP)}, pages 8107--8117, Online. Association
  for Computational Linguistics.

\bibitem[{Einolghozati et~al.(2019)Einolghozati, Gupta, Mohit, and
  Shah}]{einolghozati2019improving}
Arash Einolghozati, Sonal Gupta, Mrinal Mohit, and Rushin Shah. 2019.
\newblock Improving robustness of task oriented dialog systems.
\newblock \emph{arXiv preprint arXiv:1911.05153}.

\bibitem[{El~Asri et~al.(2017)El~Asri, Schulz, Sharma, Zumer, Harris, Fine,
  Mehrotra, and Suleman}]{asri2017frames}
Layla El~Asri, Hannes Schulz, Shikhar Sharma, Jeremie Zumer, Justin Harris,
  Emery Fine, Rahul Mehrotra, and Kaheer Suleman. 2017.
\newblock \href {https://doi.org/10.18653/v1/W17-5526} {{F}rames: a corpus for
  adding memory to goal-oriented dialogue systems}.
\newblock In \emph{Proceedings of the 18th Annual {SIG}dial Meeting on
  Discourse and Dialogue}, pages 207--219, Saarbr{\"u}cken, Germany.
  Association for Computational Linguistics.

\bibitem[{Eric et~al.(2019)Eric, Goel, Paul, Sethi, Agarwal, Gao, and
  Hakkani-Tur}]{eric2019multiwoz}
Mihail Eric, Rahul Goel, Shachi Paul, Abhishek Sethi, Sanchit Agarwal, Shuyag
  Gao, and Dilek Hakkani-Tur. 2019.
\newblock Multi{WOZ} 2.1: Multi-domain dialogue state corrections and state
  tracking baselines.
\newblock \emph{arXiv preprint arXiv:1907.01669}.

\bibitem[{Eric et~al.(2017)Eric, Krishnan, Charette, and Manning}]{eric2017key}
Mihail Eric, Lakshmi Krishnan, Francois Charette, and Christopher~D. Manning.
  2017.
\newblock \href {https://doi.org/10.18653/v1/W17-5506} {Key-value retrieval
  networks for task-oriented dialogue}.
\newblock In \emph{Proceedings of the 18th Annual {SIG}dial Meeting on
  Discourse and Dialogue}, pages 37--49, Saarbr{\"u}cken, Germany. Association
  for Computational Linguistics.

\bibitem[{Ganhotra et~al.(2020)Ganhotra, Moore, Joshi, and
  Wadhawan}]{ganhotra-etal-2020-effects}
Jatin Ganhotra, Robert Moore, Sachindra Joshi, and Kahini Wadhawan. 2020.
\newblock \href {https://www.aclweb.org/anthology/2020.findings-emnlp.358}
  {Effects of naturalistic variation in goal-oriented dialog}.
\newblock In \emph{Findings of the Association for Computational Linguistics:
  EMNLP 2020}, pages 4013--4020, Online. Association for Computational
  Linguistics.

\bibitem[{Han et~al.(2020)Han, Liu, Takanobu, Lian, Huang, Peng, and
  Huang}]{han2020multiwoz}
Ting Han, Ximing Liu, Ryuichi Takanobu, Yixin Lian, Chongxuan Huang, Wei Peng,
  and Minlie Huang. 2020.
\newblock Multiwoz 2.3: A multi-domain task-oriented dataset enhanced with
  annotation corrections and co-reference annotation.
\newblock \emph{arXiv preprint arXiv:2010.05594}.

\bibitem[{Heck et~al.(2020)Heck, van Niekerk, Lubis, Geishauser, Lin, Moresi,
  and Gasic}]{heck2020trippy}
Michael Heck, Carel van Niekerk, Nurul Lubis, Christian Geishauser, Hsien-Chin
  Lin, Marco Moresi, and Milica Gasic. 2020.
\newblock \href {https://aclanthology.org/2020.sigdial-1.4} {{T}rip{P}y: A
  triple copy strategy for value independent neural dialog state tracking}.
\newblock In \emph{Proceedings of the 21th Annual Meeting of the Special
  Interest Group on Discourse and Dialogue}, pages 35--44, 1st virtual meeting.
  Association for Computational Linguistics.

\bibitem[{Hosseini-Asl et~al.(2020)Hosseini-Asl, McCann, Wu, Yavuz, and
  Socher}]{hosseini2020simple}
Ehsan Hosseini-Asl, Bryan McCann, Chien-Sheng Wu, Semih Yavuz, and Richard
  Socher. 2020.
\newblock \href
  {https://proceedings.neurips.cc/paper/2020/file/e946209592563be0f01c844ab2170f0c-Paper.pdf}
  {A simple language model for task-oriented dialogue}.
\newblock In \emph{Advances in Neural Information Processing Systems},
  volume~33, pages 20179--20191. Curran Associates, Inc.

\bibitem[{Hu et~al.(2019)Hu, Singh, Holzenberger, Post, and
  Van~Durme}]{parabank2}
J.~Edward Hu, Abhinav Singh, Nils Holzenberger, Matt Post, and Benjamin
  Van~Durme. 2019.
\newblock \href {https://doi.org/10.18653/v1/K19-1005} {Large-scale, diverse,
  paraphrastic bitexts via sampling and clustering}.
\newblock In \emph{Proceedings of the 23rd Conference on Computational Natural
  Language Learning (CoNLL)}, pages 44--54, Hong Kong, China. Association for
  Computational Linguistics.

\bibitem[{Kelley(1984)}]{kelley1984iterative}
John~F Kelley. 1984.
\newblock An iterative design methodology for user-friendly natural language
  office information applications.
\newblock \emph{ACM Transactions on Information Systems (TOIS)}, 2(1):26--41.

\bibitem[{Lee et~al.(2019)Lee, Lee, and Kim}]{Lee2019Jul}
Hwaran Lee, Jinsik Lee, and Tae-Yoon Kim. 2019.
\newblock \href {https://doi.org/10.18653/v1/p19-1546} {Sumbt: Slot-utterance
  matching for universal and scalable belief tracking}.
\newblock \emph{Proceedings of the 57th Annual Meeting of the Association for
  Computational Linguistics}.

\bibitem[{Lewis et~al.(2020)Lewis, Liu, Goyal, Ghazvininejad, Mohamed, Levy,
  Stoyanov, and Zettlemoyer}]{lewis2019bart}
Mike Lewis, Yinhan Liu, Naman Goyal, Marjan Ghazvininejad, Abdelrahman Mohamed,
  Omer Levy, Veselin Stoyanov, and Luke Zettlemoyer. 2020.
\newblock \href {https://doi.org/10.18653/v1/2020.acl-main.703} {{BART}:
  Denoising sequence-to-sequence pre-training for natural language generation,
  translation, and comprehension}.
\newblock In \emph{Proceedings of the 58th Annual Meeting of the Association
  for Computational Linguistics}, pages 7871--7880, Online. Association for
  Computational Linguistics.

\bibitem[{McClosky et~al.(2006)McClosky, Charniak, and
  Johnson}]{mcclosky2006effective}
David McClosky, Eugene Charniak, and Mark Johnson. 2006.
\newblock \href {https://aclanthology.org/N06-1020} {Effective self-training
  for parsing}.
\newblock In \emph{Proceedings of the Human Language Technology Conference of
  the {NAACL}, Main Conference}, pages 152--159, New York City, USA.
  Association for Computational Linguistics.

\bibitem[{Moradshahi et~al.(2020)Moradshahi, Campagna, Semnani, Xu, and
  Lam}]{moradshahi-etal-2020-localizing}
Mehrad Moradshahi, Giovanni Campagna, Sina Semnani, Silei Xu, and Monica Lam.
  2020.
\newblock \href {https://www.aclweb.org/anthology/2020.emnlp-main.481}
  {Localizing open-ontology {QA} semantic parsers in a day using machine
  translation}.
\newblock In \emph{Proceedings of the 2020 Conference on Empirical Methods in
  Natural Language Processing (EMNLP)}, pages 5970--5983, Online. Association
  for Computational Linguistics.

\bibitem[{Peng et~al.(2020)Peng, Li, Li, Shayandeh, Liden, and
  Gao}]{peng2020soloist}
Baolin Peng, Chunyuan Li, Jinchao Li, Shahin Shayandeh, Lars Liden, and
  Jianfeng Gao. 2020.
\newblock Soloist: Few-shot task-oriented dialog with a single pre-trained
  auto-regressive model.
\newblock \emph{arXiv preprint arXiv:2005.05298}.

\bibitem[{Rastogi et~al.(2020)Rastogi, Zang, Sunkara, Gupta, and
  Khaitan}]{rastogi2019towards}
Abhinav Rastogi, Xiaoxue Zang, Srinivas Sunkara, Raghav Gupta, and Pranav
  Khaitan. 2020.
\newblock Towards scalable multi-domain conversational agents: The
  schema-guided dialogue dataset.
\newblock In \emph{Proceedings of the AAAI Conference on Artificial
  Intelligence}, volume~34, pages 8689--8696.

\bibitem[{Ren et~al.(2019)Ren, Ni, and McAuley}]{ren2019scalable}
Liliang Ren, Jianmo Ni, and Julian McAuley. 2019.
\newblock \href {https://doi.org/10.18653/v1/d19-1196} {Scalable and accurate
  dialogue state tracking via hierarchical sequence generation}.
\newblock \emph{Proceedings of the 2019 Conference on Empirical Methods in
  Natural Language Processing and the 9th International Joint Conference on
  Natural Language Processing (EMNLP-IJCNLP)}.

\bibitem[{Shah et~al.(2018)Shah, Hakkani-T{\"u}r, T{\"u}r, Rastogi, Bapna,
  Nayak, and Heck}]{DBLP:journals/corr/abs-1801-04871}
Pararth Shah, Dilek Hakkani-T{\"u}r, Gokhan T{\"u}r, Abhinav Rastogi, Ankur
  Bapna, Neha Nayak, and Larry Heck. 2018.
\newblock Building a conversational agent overnight with dialogue self-play.
\newblock \emph{arXiv preprint arXiv:1801.04871}.

\bibitem[{Tellman(2021)}]{smdataflow2021}
Zachary Tellman. 2021.
\newblock Designing a framework for conversational interfaces.
\newblock
  \url{https://www.microsoft.com/en-us/research/group/msai/articles/designing-a-framework-for-conversational-interfaces/}.

\bibitem[{Vaswani et~al.(2017)Vaswani, Shazeer, Parmar, Uszkoreit, Jones,
  Gomez, Kaiser, and Polosukhin}]{vaswani2017attention}
Ashish Vaswani, Noam Shazeer, Niki Parmar, Jakob Uszkoreit, Llion Jones,
  Aidan~N Gomez, {\L}ukasz Kaiser, and Illia Polosukhin. 2017.
\newblock Attention is all you need.
\newblock In \emph{Advances in Neural Information Processing Systems}, pages
  5998--6008.

\bibitem[{Wang et~al.(2015)Wang, Berant, and Liang}]{overnight}
Yushi Wang, Jonathan Berant, and Percy Liang. 2015.
\newblock \href {https://doi.org/10.3115/v1/p15-1129} {Building a semantic
  parser overnight}.
\newblock In \emph{Proceedings of the 53rd Annual Meeting of the Association
  for Computational Linguistics and the 7th International Joint Conference on
  Natural Language Processing (Volume 1: Long Papers)}, pages 1332--1342.
  Association for Computational Linguistics.

\bibitem[{Williams and Zweig(2016)}]{williams2016end}
Jason~D Williams and Geoffrey Zweig. 2016.
\newblock End-to-end lstm-based dialog control optimized with supervised and
  reinforcement learning.
\newblock \emph{arXiv preprint arXiv:1606.01269}.

\bibitem[{Wolf et~al.(2020)Wolf, Debut, Sanh, Chaumond, Delangue, Moi, Cistac,
  Rault, Louf, Funtowicz, Davison, Shleifer, von Platen, Ma, Jernite, Plu, Xu,
  Le~Scao, Gugger, Drame, Lhoest, and Rush}]{Wolf2019HuggingFacesTS}
Thomas Wolf, Lysandre Debut, Victor Sanh, Julien Chaumond, Clement Delangue,
  Anthony Moi, Pierric Cistac, Tim Rault, Remi Louf, Morgan Funtowicz, Joe
  Davison, Sam Shleifer, Patrick von Platen, Clara Ma, Yacine Jernite, Julien
  Plu, Canwen Xu, Teven Le~Scao, Sylvain Gugger, Mariama Drame, Quentin Lhoest,
  and Alexander Rush. 2020.
\newblock \href {https://doi.org/10.18653/v1/2020.emnlp-demos.6} {Transformers:
  State-of-the-art natural language processing}.
\newblock In \emph{Proceedings of the 2020 Conference on Empirical Methods in
  Natural Language Processing: System Demonstrations}, pages 38--45, Online.
  Association for Computational Linguistics.

\bibitem[{Wu et~al.(2019)Wu, Madotto, Hosseini-Asl, Xiong, Socher, and
  Fung}]{Wu2019May}
Chien-Sheng Wu, Andrea Madotto, Ehsan Hosseini-Asl, Caiming Xiong, Richard
  Socher, and Pascale Fung. 2019.
\newblock Transferable multi-domain state generator for task-oriented dialogue
  systems.
\newblock In \emph{Proceedings of the 57th Annual Meeting of the Association
  for Computational Linguistics}, pages 808--819.

\bibitem[{Xu et~al.(2020)Xu, Semnani, Campagna, and Lam}]{xu2020autoqa}
Silei Xu, Sina Semnani, Giovanni Campagna, and Monica Lam. 2020.
\newblock \href {https://www.aclweb.org/anthology/2020.emnlp-main.31}
  {{A}uto{QA}: From databases to {Q}{\&}{A} semantic parsers with only
  synthetic training data}.
\newblock In \emph{Proceedings of the 2020 Conference on Empirical Methods in
  Natural Language Processing (EMNLP)}, pages 422--434, Online. Association for
  Computational Linguistics.

\bibitem[{Ye et~al.(2021{\natexlab{a}})Ye, Manotumruksa, and
  Yilmaz}]{ye2021multiwoz}
Fanghua Ye, Jarana Manotumruksa, and Emine Yilmaz. 2021{\natexlab{a}}.
\newblock Multiwoz 2.4: A multi-domain task-oriented dialogue dataset with
  essential annotation corrections to improve state tracking evaluation.
\newblock \emph{arXiv preprint arXiv:2104.00773}.

\bibitem[{Ye et~al.(2021{\natexlab{b}})Ye, Manotumruksa, Zhang, Li, and
  Yilmaz}]{ye2021star}
Fanghua Ye, Jarana Manotumruksa, Qiang Zhang, Shenghui Li, and Emine Yilmaz.
  2021{\natexlab{b}}.
\newblock \href {https://doi.org/10.1145/3442381.3449939} {Slot self-attentive
  dialogue state tracking}.
\newblock In \emph{Proceedings of the Web Conference 2021}, WWW '21, page
  1598–1608, New York, NY, USA. Association for Computing Machinery.

\bibitem[{Yu et~al.(2021)Yu, Zhang, Polozov, Meek, and Awadallah}]{yu2020score}
Tao Yu, Rui Zhang, Oleksandr Polozov, Christopher Meek, and Ahmed~Hassan
  Awadallah. 2021.
\newblock \href {https://openreview.net/forum?id=oyZxhRI2RiE} {{SCoRE}:
  Pre-training for context representation in conversational semantic parsing}.
\newblock In \emph{International Conference on Learning Representations}.

\bibitem[{Zang et~al.(2020)Zang, Rastogi, Sunkara, Gupta, Zhang, and
  Chen}]{zang2020multiwoz}
Xiaoxue Zang, Abhinav Rastogi, Srinivas Sunkara, Raghav Gupta, Jianguo Zhang,
  and Jindong Chen. 2020.
\newblock \href {https://doi.org/10.18653/v1/2020.nlp4convai-1.13}
  {{M}ulti{WOZ} 2.2 : A dialogue dataset with additional annotation corrections
  and state tracking baselines}.
\newblock In \emph{Proceedings of the 2nd Workshop on Natural Language
  Processing for Conversational AI}, pages 109--117, Online. Association for
  Computational Linguistics.

\bibitem[{Zhang et~al.(2020)Zhang, Ou, and Yu}]{zhang2019task}
Yichi Zhang, Zhijian Ou, and Zhou Yu. 2020.
\newblock \href {https://doi.org/10.1609/aaai.v34i05.6507} {Task-oriented
  dialog systems that consider multiple appropriate responses under the same
  context}.
\newblock \emph{Proceedings of the AAAI Conference on Artificial Intelligence},
  34(05):9604–9611.

\bibitem[{Zhou and Small(2019)}]{zhou2019multi}
Li~Zhou and Kevin Small. 2019.
\newblock Multi-domain dialogue state tracking as dynamic knowledge graph
  enhanced question answering.
\newblock \emph{arXiv preprint arXiv:1911.06192}.

\bibitem[{Zoph et~al.(2020)Zoph, Ghiasi, Lin, Cui, Liu, Cubuk, and
  Le}]{zoph2020rethinking}
Barret Zoph, Golnaz Ghiasi, Tsung-Yi Lin, Yin Cui, Hanxiao Liu, Ekin~Dogus
  Cubuk, and Quoc Le. 2020.
\newblock Rethinking pre-training and self-training.
\newblock \emph{Advances in neural information processing systems},
  33:3833--3845.

\end{thebibliography}

\clearpage
\appendix

\section{ThingTalk Definition}
\label{sec:syntax}
\subsection{Syntax}

{
\small
\begin{tabbing}
123456789012345678\=1\=\kill
Formal Dialogue $d$ \>$:$\> $a~~\textit{r}^*~~s^*$\\
User State $u$ \>$:$\> $\textit{ua}~~s^*$\\
Agent State $a$ \>$:$\> $\textit{aa}~~\textit{as}^?$\\
User Act \textit{ua}
\>$:$    \>$\text{Greet}~~\vert~~\text{Exec}~~\vert~~\text{Cancel}~~\vert~~\text{Insist}$\\
\>$\vert$\>$\text{AskRecommend}~~\vert~~\text{LearnMore}$\\
\>$\vert$\>$\text{ActionQuestion}~~\vert~~\text{End}~~\vert~~\text{Invalid}$\\
Agent Act \textit{aa} \>$:$\>$\text{Init}~~\vert~~\text{Greet}~~\vert~~\text{RecommendOne}$\\
                      \>$\vert$\>$\text{RecommendMany}~~\vert~~\text{Propose}$\\
                      \>$\vert$\>$\text{SearchQuestion}~~\vert~~\text{SlotFill}$\\
                      \>$\vert$\>$\text{LearnMoreWhat}~~\vert~~\text{EmptySearch}$\\
                      \>$\vert$\>$\text{Confirm}~~\vert~~\text{ActionSuccess}$\\
                      \>$\vert$\>$\text{ActionError}~~\vert~~\text{AnythingElse}$\\
                      \>$\vert$\>$\text{Invalid}$\\
User Statement $s$ \>$:$\> $q~~\vert~~\textit{ac}$ \\
Result $r$ \>$:$\> $s~~\left[ \texttt{\{} [\textit{sn} = v]^+ \texttt{\}}\right]*$ \\
Agent Statement \textit{as} \>$:$\>$\text{Request}~~ \textit{sn}^+$\\
\>$\vert$\>$[\text{Propose}~~\vert~~\text{Confirm}]~~[q~~\vert~~\textit{ac}]$  \\
Query $q$ \>$:$\> <ThingTalk query>\\
Action $\textit{ac}$ \>$:$\> $\textit{dn} \texttt{(}[\textit{sn} = v]^*\texttt{)}$\\
Domain Name \textit{dn} \>$:$\> <identifier>\\
Slot Name \textit{sn} \>$:$\> <identifier>\\
Value $v$ \>$:$\> <constant>
\end{tabbing}
}

\subsection{Agent Definition}
The agent is a function $\text{Agent}(d, u) = d'$ that computes the new formal representation of the entire dialogue. The representation is constructed incrementally, starting from the initial dialogue $d_0$ which is empty.

Let $d = (a, r, s) \in \D$ and $u = (\textit{ua}, s_\text{u}) \in U$ be the two inputs to the agent. The agent computes the new agent state $a$ as follows:
\begin{align*}
    (r_\text{u}, \textit{is}_\text{u}) &=  \text{Execute}(s_\text{u}) \\
    a' &= \text{Policy}(\textit{ua}, r \vert\vert r_\text{u}, \textit{is}_\text{u}) \\
    d' &= (a', r \vert\vert r_\text{u}, \textit{is}_\text{u})
\end{align*}
where $\vert\vert$ denotes concatenation. The Execute function calls the ThingTalk runtime to execute the statements in the user state, $s_\text{u}$. It returns (1) the results $r_\text{u}$ by executing all statements in $s_\text{u}$ whose required parameters are available, (2) the rest of the (incomplete) statements, $\textit{is}_\text{u}$.  The Policy function determines the agent state $a'$ from the user state $\textit{ua}$. all the results $r_\text{u}$ appended to previous results $r$, and $\textit{is}_\text{u}$.  
The agent returns the new dialogue $d'$ with the new agent state, all the results and the new incomplete statements. 
The incomplete statements $s$ in $d$ are discarded. If the user has not changed topics, information in $s$ is incorporated in $s_\text{u}$.

\section{Training}
\label{sec:hyperparameter}
\subsection{Preprocessing}
We apply the same preprocessing used by TRADE~\cite{Wu2019May} to the input utterances. We also use a rule-based preprocessor to identify time expressions, and replace them with placeholder tokens. All slot values in the result and agent states that have string or time type are replaced with a placeholder when input to the model.

We normalize all slot values in the user state to match the utterance, regardless of typos.
When comparing the slot values for equality, we normalize entity names via a database lookup.

\subsection{Hyperparameters}

Our model uses a BART large model which has 400 million trainable parameters. We use the Adam optimizer, with the Transformer learning rate schedule (800 iterations of warm-up, 0.04 multiplier).

We train our model for 50,000 gradient updates on the synthesized data and choose the model with the highest validation exact-match accuracy. We then fine-tune that model on the few-shot training set for 15,000 gradient updates, and again choose the model with the highest validation accuracy. We repeat this process for another 15,000 updates on the self-train set. Training is done on a single GPU with 16 GB memory and batch size is chosen based on the length of the examples in each batch: we choose as many examples as we can fit in the GPU memory. Gradient accumulation is used to increase the effective batch size by a factor of 20.

\section{Dataset Statistics}
\label{sec:data-stats}
\begin{table}[H]

\small
\centering
\begin{tabular}{lrrr}
\toprule
     & {\bf \# dlgs} & {\bf \# turns} & \bf{\# words} \\
\midrule
    Synthesized & 968,007  & 830,792   & 11,390,957  \\
    Paraphrased & 592,970 & 945,946   & 12,845,548  \\
    Few-Shot & 168 & 1,061 & 14,669 \\
    Self-Training & 8,420 & 56,546 & 760,927 \\
\midrule
    Validation & 265 & 1,582 & 21,256  \\
    Test & 995  & 7,271 & 100,814 \\
\bottomrule
\end{tabular}

\caption{Statistics of our training and evaluation sets: number of dialogues, of turns, and of words. For the synthesized dataset, we do not count turns that appear identically in multiple dialogues.}
\label{table:dataset}
\end{table}

\section{Annotated Example Dialogue}

Here is an example of a dialogue generated by the simulator, between the user U and the agent A. US denotes the user state, D denotes the formal dialogue.

\begin{figure}[H]
\include{example}
\end{figure}

\end{document}